\def\year{2020}\relax
\documentclass[letterpaper]{article} 
\usepackage{aaai20}  
\usepackage{times}  
\usepackage{helvet} 
\usepackage{courier}  
\usepackage[hyphens]{url}  
\usepackage{graphicx} 
\usepackage{graphicx}  
\usepackage{amssymb}
\usepackage{amsmath}
\usepackage{graphicx}
\usepackage{subcaption}
\usepackage{booktabs} 
\usepackage{multirow}
\usepackage{diagbox}

\makeatletter
\def\@biblabel#1{}
\makeatother

\title{Structured Output Learning with Conditional Generative Flows}
\author{You Lu\\ Department of Computer Science \\
	Virginia Tech \\
	Blacksburg, VA\\
	\tt you.lu@vt.edu
	\And
	Bert Huang\\ Department of Computer Science \\
	Virginia Tech \\
	Blacksburg, VA \\
	{\tt bhuang@vt.edu}
}
\begin{document}
	
	\maketitle
	
	\begin{abstract}
		Traditional structured prediction models try to learn the conditional likelihood, i.e., $p(y|x)$, to capture the relationship between the structured output $y$ and the input features $x$. For many models, computing the likelihood is intractable. These models are therefore hard to train, requiring the use of surrogate objectives or variational inference to approximate likelihood. In this paper, we propose conditional Glow (c-Glow), a conditional generative flow for structured output learning. C-Glow benefits from the ability of flow-based models to compute $p(y|x)$ exactly and efficiently. Learning with c-Glow does not require a surrogate objective or performing inference during training. Once trained, we can directly and efficiently generate conditional samples. We develop a sample-based prediction method, which can use this advantage to do efficient and effective inference. In our experiments, we test c-Glow on five different tasks. C-Glow outperforms the state-of-the-art baselines in some tasks and predicts comparable outputs in the other tasks. The results show that c-Glow is versatile and is applicable to many different structured prediction problems.
	\end{abstract}
	
	\section{Introduction}
	\label{sec:introduction}
	
	Structured prediction models  are widely used in tasks such as image segmentation~\cite{nowozin2011structured} and sequence labeling~\cite{lafferty2001conditional}. In these structured output tasks, the goal is to model a mapping from the input $x$ to the high-dimensional structured output $y$. In many such problems, it is also important to make diverse predictions to capture the variability of plausible solutions to the structured output problem~\cite{sohn2015learning}. 
	
	Many existing methods for structured output learning use graphical models, such as conditional random fields (CRFs) \cite{wainwright2008graphical}, and approximate the conditional distribution $p(y|x)$. Approximation is necessary because, for most graphical models, computing the exact likelihood is intractable. Recently, deep structured prediction models~\cite{chen2015learning,zheng2015conditional,sohn2015learning,wang2016proximal,belanger2016structured,graber2018deep} combine deep neural networks with graphical models, using the power of deep neural networks to extract high-quality features and graphical models to model correlations and dependencies among variables. The main drawback of these approaches is that, due to the intractable likelihood, they are difficult to train. Training them requires the construction of surrogate objectives, or approximating the likelihood by using variational inference to infer latent variables. Moreover, once the model is trained, inference and sampling from CRFs require expensive iterative procedures \cite{koller2009probabilistic}.
	
	In this paper, we develop conditional generative flows (c-Glow) for structured output learning. Our model is a variant of Glow~\cite{kingma2018glow}, with additional neural networks for capturing the relationship between input features and structured output variables. Compared to most methods for structured output learning, c-Glow has the unique advantage that it can directly model the conditional distribution $p(y|x)$ without restrictive assumptions (e.g., variables being fully connected \cite{krahenbuhl2011efficient}). We can train c-Glow by exploiting the fact that invertible flows allow exact computation of log-likelihood, removing the need for surrogates or inference.  Compared to other methods using normalizing flows (e.g., \cite{trippe2018conditional,kingma2018glow}), c-Glow's output label $y$ is conditioned on both complex input and a high-dimensional tensor rather than a one-dimensional scalar. We evaluate c-Glow on five structured prediction tasks: binary segmentation, multi-class segmentation, color image denoising, depth refinement, and image inpainting, finding that c-Glow's exact likelihood training is able to learn models that efficiently predict structured outputs of comparable quality to state-of-the-art deep structured prediction approaches.
	
	\section{Related Work}
	\label{sec:relatedwork}
	
	There are two main topics of research related to our paper: deep structured prediction and normalizing flows. In this section, we briefly cover some of the most related literature.
	
	\subsection{Deep Structured Models}
	
	One emerging strategy to construct deep structured models is to combine deep neural networks with graphical models. However, this kind of model can be difficult to train, since the likelihood of graphical models is usually intractable. \citeauthor{chen2015learning} (\citeyear{chen2015learning}) proposed joint learning approaches that blend the learning and approximate inference to alleviate some of these computational challenges.  \citeauthor{zheng2015conditional} (\citeyear{zheng2015conditional}) proposed CRF-RNN, a method that treats mean-field variational CRF inference as a recurrent neural network to allow gradient-based learning of model parameters. \citeauthor{wang2016proximal} (\citeyear{wang2016proximal}) proposed proximal methods for inference. And \citeauthor{sohn2015learning} (\citeyear{sohn2015learning}) used variational autoencoders~\cite{kingma2013auto} to generate latent variables for predicting the output. 
	While using a surrogate for the true likelihood is generally viewed as a concession, \citeauthor{norouzi2016reward} (\citeyear{norouzi2016reward}) found that training with a tractable task-specific loss often yielded better performance for the goal of reducing specific task losses than training with general-purpose likelihood approximations. Their analysis hints that fitting a distribution with a true likelihood may not always train the best predictor for specific tasks.
	
	Another direction combining structured output learning with deep models is to construct energy functions with deep networks. Structured prediction energy networks (SPENs) \cite{belanger2016structured} define energy functions for scoring structured outputs as differentiable deep networks. The likelihood of a SPEN is intractable, so the authors used structured SVM loss to learn. SPENs can also be trained in an end-to-end learning framework \cite{belanger2017end} based on unrolled optimization. Methods to alleviate the cost of SPEN inference include replacing the argmax inference with an inference network \cite{tu2018learning}. Inspired by Q-learning, \citeauthor{gygli2017deep} (\citeyear{gygli2017deep}) used an oracle value function as the objective for  energy-based deep networks. \citeauthor{graber2018deep} (\citeyear{graber2018deep}) generalized SPENs by adding non-linear transformations on top of the score function.
	
	\subsection{Normalizing Flows}
	
	Normalizing flows are neural networks constructed with fully invertible components. The invertibility of the resulting network provides various mathematical benefits.
	Normalizing flows have been successfully used to build likelihood-based deep generative models~\cite{dinh2014nice,dinh2016density,kingma2018glow} and to improve variational approximation~\cite{rezende2015variational,kingma2016improved}. Autoregressive flows~\cite{kingma2016improved,papamakarios2017masked,huang2018neural,ziegler2019latent} condition each affine transformation on all previous variables, so that they ensure an invertible transformation and triangular Jacobian matrix. Continuous normalizing flows~\cite{chen2018neural,grathwohl2018ffjord} define the transformation function using  ordinary differential equations. While most normalizing flow models define generative models, \citeauthor{trippe2018conditional} (\citeyear{trippe2018conditional}) developed radial flows to model univariate conditional probabilities. 
	
	Most related to our approach are flow-based generative models for complex output. \citeauthor{dinh2014nice} (\citeyear{dinh2014nice}) first proposed a flow-based model, NICE, for modeling complex high-dimensional densities. They later proposed Real-NVP~\cite{dinh2016density}, which improves the expressiveness of NICE by adding more flexible coupling layers. The Glow model~\cite{kingma2018glow} further improved the performance of such approaches by incorporating new invertible layers. Most recently, Flow++~\cite{ho2019flow} improved generative flows with variational dequantization and architecture design, and \citeauthor{hoogeboom2019emerging} (\citeyear{hoogeboom2019emerging}) proposed new invertible convolutional layers for flow-based models.
	
	\section{Background}
	\label{sec:background}
	
	In this section, we introduce notation and background knowledge directly related to our work.
	
	\subsection{Structured Output Learning}
	
	Let $x$ and $y$ be random variables with unknown true distribution $p^*(y|x)$. We collect a dataset $\mathcal{D}=\{(x_1,y_1), ...,(x_N, y_N)\}$, where $x_i$ is the $i$th input and $y_i$ is the corresponding output. We approximate $p^*(y|x)$ with a model $p(y|x,\theta)$ and minimize the negative log-likelihood
	\begin{equation*}
	\mathcal{L(D)} = - \frac{1}{N} \sum_{i=1}^{N} \log p(y_i| x_i, \theta).
	\end{equation*}
	
	In structured output learning, the label $y$ comes from a complex, high-dimensional output space $\mathcal{Y}$ with dependencies among output dimensions. Many structured output learning approaches use an energy-based model to define a conditional distribution:
	\begin{equation*}
	p(y|x) = \frac{e^{E(y,x)}}{\int_{y' \in \mathcal{Y}}e^{E(y',x)} dy},
	\end{equation*}
	where $E(.,.): \mathcal{X}\times \mathcal{Y} \to \mathcal{R}$ is the energy function. In deep structured prediction, $E(x,y)$ depends on $x$ via a deep network. Due to the high dimensionality of $y$, the partition function, i.e., $\int_{y' \in \mathcal{Y}}e^{E(y',x)} dy$, is intractable. To train the model, we need methods to approximate the partition function such as variational inference or surrogate objectives, resulting in complicated training and sub-optimal results. 
	
	\subsection{Conditional Normalizing Flows}
	
	A normalizing flow is a composition of invertible functions $f = f_1 \circ f_2 \circ \dots \circ f_M$, which transforms the target $y$ to a latent code $z$ drawn from a simple distribution. In conditional normalizing flows~\cite{trippe2018conditional}, we rewrite each function as $f_i = f_{x, \phi_i}$, making it parameterized by both $x$ and its parameter $\phi_i$. Thus, with the \emph{change of variables} formula, we can rewrite the conditional likelihood as
	\begin{equation}
	\log p(y|x,\theta) = \log p_Z(z) + \sum_{i=1}^{M} \log \left|\det \left(\frac{\partial f_{x, \phi_i}}{\partial r_{i-1}}\right)\right|,
	\label{eq:cflow}
	\end{equation}
	where $r_i = f_{\phi_i}(r_{i-1})$, $r_{0} = x$, and $r_{M}=z$. 
	
	In this paper, we address the structured output problem by using normalizing flows. That is, we directly use the conditional normalizing flows, i.e., Equation~\ref{eq:cflow}, to calculate the conditional distribution. Thus, the model can be trained by locally optimizing the exact likelihood. Note that conditional normalizing flows have been used for conditional density estimation. \citeauthor{trippe2018conditional} (\citeyear{trippe2018conditional}) use it to solve the one-dimensional regression problem. Our method is different from theirs in that the labels in our problem are high-dimensional tensors rather than scalars. We therefore will build on recently developed methods for (unconditional) flow-based generative models for high-dimensional data.
	
	\subsection{Glow}
	
	Glow~\cite{kingma2018glow} is a flow-based generative model that extends other flow-based models: NICE~\cite{dinh2014nice} and Real-NVP~\cite{dinh2016density}. Glow's modifications have demonstrated significant improvements in likelihood and sample quality for natural images. The model mainly consists of three components. Let $u$ and $v$ be the input and output of a layer, whose shape is $[h \times w\times c]$, with spatial dimensions $(h,w)$ and channel dimension $c$. The three components are as follows.
	
	\noindent\textbf{Actnorm layers.} Each activation normalization (actnorm) layer performs an affine transformation of activations using two $1 \times c$ parameters, i.e., a scalar $s$,  and a bias $b$. The transformation can be written as
	\begin{equation*}
	u_{i,j} = s \odot v_{i,j} + b,
	\end{equation*}
	where $\odot$ is the element-wise product.
	
	\noindent\textbf{Invertible 1$\times$1 convolutional layers.} Each invertible 1x1 convolutional layer is a generalization of a permutation operation. Its function format is
	\begin{equation*}
	u_{i,j} = Wv_{i,j},
	\end{equation*}
	where $W$ is a $c \times c$ weight matrix.
	
	\noindent\textbf{Affine layers.} As in the NICE and Real-NVP models, Glow also has affine coupling layers to capture the correlations among spatial dimensions. Its transformation is
	\begin{eqnarray*}
		&&v_1, v_2 = \text{split}(v),~~~~~~s_2, b_2 = \text{NN}(v_1),  \\
		&&u_2 = s_2 \odot v_2 + b_2,~~~u = \text{concat}(v_1, u_2),
	\end{eqnarray*}
	where $\text{NN}$ is a neural network, and the $\text{split}()$ and $\text{concat}()$ functions perform operations along the channel dimension. The $s_2$ and $b_2$ vectors have the same size as $v_2$.
	
	Glow uses a multi-scale architecture~\cite{dinh2016density} to combine the layers. This architecture has a ``squeeze'' layer for shuffling the variables and a ``split'' layer for reducing the computational cost. 
	
	\section{Conditional Generative Flows for Structured Output Learning}
	\label{sec:method}
	
	\begin{figure}[tp]
		\centering
		\begin{subfigure}[t]{0.23\textwidth}
			\centering
			\includegraphics[width=.8\linewidth]{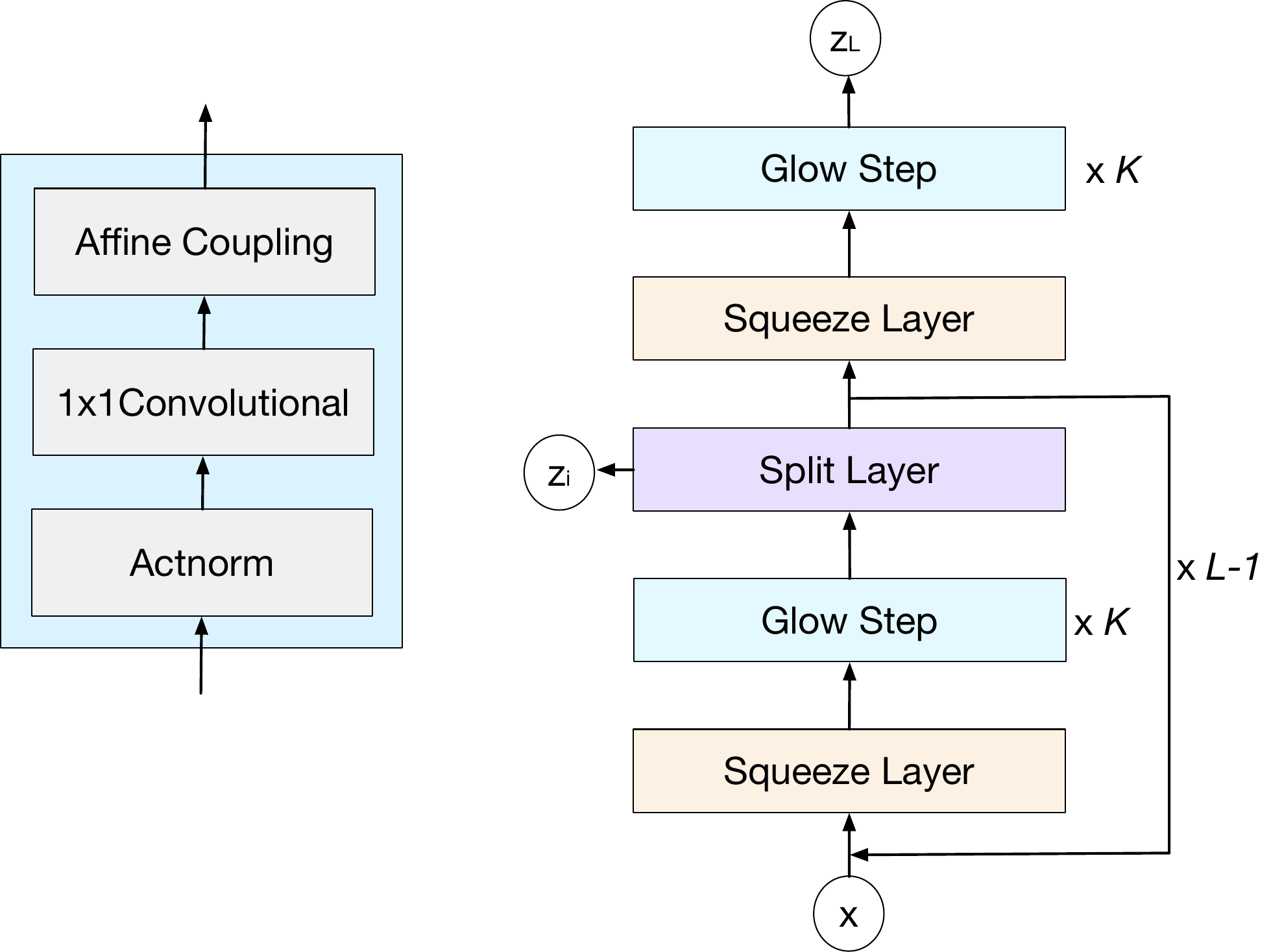}  
			\caption{Glow architecture}
			\label{fig:a}
		\end{subfigure}
		\begin{subfigure}[t]{0.23\textwidth}
			\centering
			\includegraphics[width=.9\linewidth]{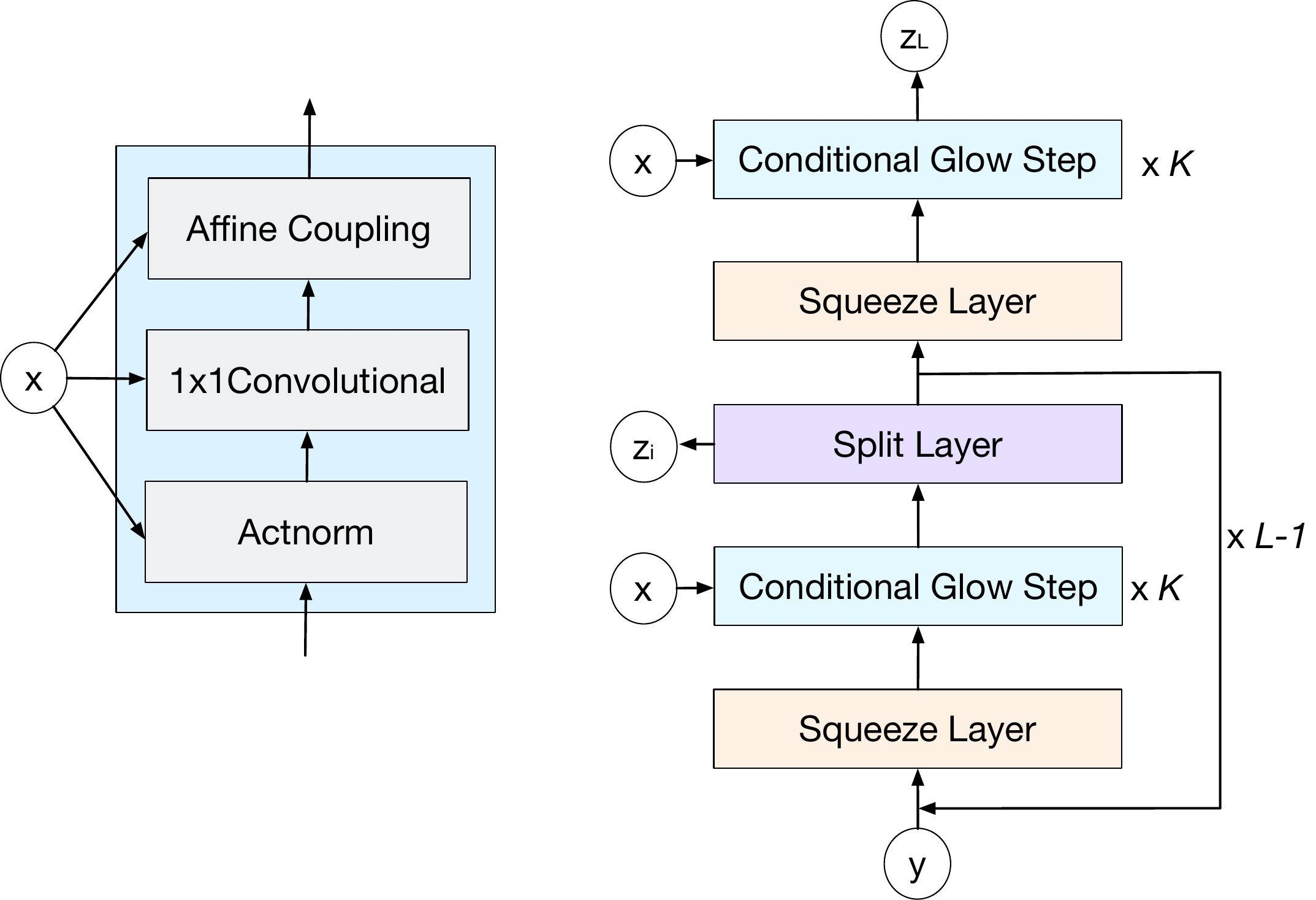}  
			\caption{c-Glow architecture}
			\label{fig:b}
		\end{subfigure}
		\caption{Model architectures for Glow and conditional Glow. For each model, the left sub-graph is the architecture of each step, and the right sub-graph is the whole architecture. The parameter $L$ represents the number of levels, and $K$ represents the depth of each level.}
		\label{fig:arc}
		\vspace{-0.1in}
	\end{figure}
	
	This section describes our conditional generative flow (c-Glow), a flow-based model for structured prediction. 
	
	\subsection{Conditional Glow}
	
	To modify Glow to be a conditional generative flow, we need to add conditioning architectures to its three components: the actnorm layer, the 1$\times$1 convolutional layer, and the affine coupling layer. The main idea is to use a neural network, which we refer to as a \emph{conditioning network} (CN), to generate the parameter weights for each layer. The details are as follows.
	
	\noindent\textbf{Conditional actnorm.} The parameters of an actnorm layer are two $1 \times c$ vectors, i.e., the scale $s$ and the bias $b$. In conditional Glow, we use a CN to generate these two vectors and then use them to transform the variable, i.e.,
	\begin{eqnarray*}
		s, b = \text{CN}(x),~~~u_{i,j} = s \odot v_{i,j} + b.
	\end{eqnarray*}
	
	\noindent\textbf{Conditional 1$\times$1 convolutional.} The 1$\times$1 convolutional layer uses a $c \times c$ weight matrix to permute each spatial dimension's variable. In conditional Glow, we use a conditioning network to generate this matrix:
	\begin{eqnarray*}
		W = \text{CN}(x) ,~~~u_{i,j} = Wv_{i,j}.
	\end{eqnarray*}
	
	\noindent\textbf{Conditional affine coupling.} The affine coupling layer separates the input variable into two halves, i.e., $v_1$ and $v_2$. It uses $v_1$ as the input to an NN to generate scale and bias parameters for $v_2$. To build a conditional affine coupling layer, we use a CN to extract features from $x$, and then we concatenate it with $v_1$ to form the input of NN.
	\begin{eqnarray*}
		&&v_1, v_2 = \text{split}(v),~~~~~~~~~x_r = \text{CN}(x) ,\\
		&&s_2, b_2 = \text{NN}(v_1, x_r),~~~u_2 = s_2 \odot v_2 + b_2, \\
		&&u = \text{concat}(v_1, u_2).
	\end{eqnarray*}
	
	We can still use the multi-scale architecture to combine these conditional components to preserve the efficiency of computation. Figure~\ref{fig:arc} illustrates the Glow and c-Glow architectures for comparison.
	
	Since the conditioning networks do not need to be invertible when optimizing a conditional model, we define the general approach without restrictions to their architectures here. Any differentiable network suffices and preserves the ability of c-Glow to compute the exact conditional likelihood of each input-output pair. We will specify the architectures we use in our experiments in Section~\ref{subsec:arc}.
	
	\subsection{Learning}
	
	To learn the model parameters, we can take advantage of the efficiently computable log-likelihood for flow-based models. In cases where the output is continuous, the likelihood calculation is direct. Therefore, we can back-propagate to differentiate the exact conditional likelihood, i.e., Eq.~\ref{eq:cflow}, and optimize all c-Glow parameters using gradient methods.
	
	In cases where the output is discrete, we follow \cite{dinh2016density,kingma2018glow,ho2019flow} and add uniform noise to $y$ during training to dequantize the data. This procedure augments the dataset and prevents model collapse. We can still use back-propagation and gradient methods to optimize the likelihood of this approximate continuous distribution. By expanding the proofs by \citeauthor{theis2015note} (\citeyear{theis2015note}) and \citeauthor{ho2019flow} (\citeyear{ho2019flow}), we can show that the discrete distribution is lower-bounded by this continuous distribution. 
	
	With a slight abuse of notation, we let $q(y|x)$ be our discrete hypothesis distribution and $p(v|x)$ be the dequantized continuous model. Then our goal is to maximize the likelihood $q$, which can be expressed by marginalizing over values of $v$ that round to $y$:
	\begin{equation*}
	q(y|x) = \int_{u \in [-0.5, 0.5)^d} p(y+u|x)du, 
	\end{equation*}
	where $d$ is the variable's dimension, and $u$ represents the difference between the continuous variable $v$ and the rounded, quantized $y$. 
	
	Let $p_d(x,y)$ be the true data distribution, and $\tilde{p}_d(x,y)$ be the distribution of the dequantized dataset. The learning process maximizes $\mathbb{E}_{\tilde{p}_{d}(x,y)}[\log p(v|x)]$. We expand this and apply Jensen's Inequality to obtain the bound:
	\begin{eqnarray*}
		&&\mathbb{E}_{\tilde{p}_{d}(x,y)}[\log p(v|x)] \\
		&=& \int_{x} \sum_{y} p_d(x,y) dx \int_{u} \log p(y+u|x)du \\
		&\le& \int_{x} \sum_{y} p_d(x,y) dx \log \int_{u} p(y+u|x)du \\
		&=& \mathbb{E}_{p_{d}(x,y)}[\log q(y|x)].
	\end{eqnarray*}
	Therefore, when $y$ is discrete, the learning optimization, which maximizes the continuous likelihood $p(v|x)$, maximizes a lower bound on $q(y|x)$.
	
	\subsection{Inference}
	
	Given a learned model $p(y|x)$, we can perform efficient sampling with a single forward pass through the c-Glow. We first calculate the transformation functions given $x$ and then sample the latent code $z$ from $p_Z(z)$. Finally, we propagate the sampled $z$ through the model, and we get the corresponding sample $y$. The whole process can be summarized as
	\begin{equation}
	z \sim p_Z(z), y = g_{x,\phi} (z),
	\label{eq:gen}
	\end{equation}
	where $g_{x,\phi} = f^{-1}_{x,\phi}$ is the inverse function. 
	
	The core task in structured output learning is to predict the best output, i.e., $y^*$, for an input $x$. This process can be formalized as looking for an optimized $y^*$ such that
	\begin{equation}
	y^* = \arg \max_y p(y|x).
	\label{eq:predict}
	\end{equation}
	
\begin{figure}[htp]
		\centering
		\tiny{Input Image\hspace{0.3in} Ground Truth \hspace{0.3in} Gradient-based \hspace{0.3in} Sampled-based}
		\includegraphics[width=1\linewidth]{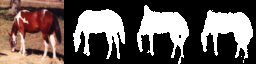}  
		\caption{Illustration of difference between a gradient-based method and a sample-based method. From left to right: the input image, the ground truth label, the gradient-based prediction, and the sample-based prediction. In the third image, the horse has a horn on its back. This is because the gradient-based method is trapped into a local optimum, which assumes the head of this horse should be in that place. In the fourth image, the sample average smooths out the horn because most samples do not have the horn mistake.}
		\label{fig:illustration}
		\vspace{-0.1in}
	\end{figure}

	To compute Equation \ref{eq:predict}, we can use gradient-based optimization, e.g., to optimize $y$ based on gradient descent. However, in our experiments, we found that this method is always slow, i.e., it takes thousands of iterations to converge. Worse, since the probability density function is non-convex with a highly multi-modal surface, it often gets stuck in local optima, resulting in sub-optimal prediction. Therefore, we use a sample-based method to approximate the inference instead. Let $\{z_1,...,z_M\}$ be samples drawn from $p_Z(z)$. Estimated marginal expectations for each variable can be computed from the average:
	\begin{equation}
	y^* \approx \frac{1}{M} \sum_{i=1}^{M} g_{x,\phi}(z_i).
	\end{equation} 
	This sample-based method can overcome the gradient-based method's problems. In our experiments, we found that we only need $10$ samples to get a high quality prediction, so inference is faster. The sample average can smooth out some anomalous values, further improving prediction. One illustration of difference between the gradient-based method and the sample-based method is in Figure~\ref{fig:illustration}.
	
	When $y$ is a continuous variable, we can directly get $y^*$ from the above sample-based prediction. When $y$ is discrete, we follow previous literature~\cite{belanger2016structured,gygli2017deep} to round $y^*$ to discrete values. In our experiments, we find that the predicted $y^*$ values are already near integral values. 
	
	\section{Experiments}
	\label{sec:experiments}
	
	In this section, we evaluate c-Glow on five structured prediction tasks: binary segmentation, multi-class segmentation, image denoising, depth refinement, and image inpainting. We find c-Glow is among the class of state-of-the-art methods while retaining its likelihood and sampling benefits.
	
	\subsection{Architecture and Setup}
	\label{subsec:arc}
	To specify a c-Glow architecture, we need to define conditioning networks that generate weights for the conditional actnorm, 1$\times$1 convolutional, and affine layers. 
	
	For the conditional actnorm layer, we use a six-layer conditioning network. The first three layers are convolutional layers that downscale the input $x$ to a reasonable size. The last three layers are then fully connected layers, which transform the resized $x$ to the scale $s$ and the bias $b$ vectors. For the downscaling convolutional layers, we use a simple method to determine their kernel size and stride. Let $H_i$ and $H_o$ be the input and output sizes. Then we set the stride to $H_i / H_o$ and the kernel size to $2 \times \text{padding} + \text{stride}$.
	
	For the conditional 1$\times$1 convolutional layer, we use a similar six-layer network to generate the weight matrix. The only difference is that the last fully connected layer will generate the weight matrix $W$. For the actnorm and 1$\times$1 convolutional conditional networks, the number of channels of the convolutional layers, i.e., $n_c$, and the width of the fully connected layers, i.e., $n_w$, will impact the model's performance.
	
	For the conditional affine layer, we use a three-layer conditional network to extract features from $x$, and we concatenate it with $v_1$. Among the three layers, the first and the last layers use $3 \times 3$ kernels. The middle layer is a downscaling convolutional layer. We vary the number of channels of this conditional network to be $\{8, 16, 32\}$, and we find that the model is not very sensitive to this variation. In our experiments, we fix it to have $16$ channels. The affine layer itself is composed of three convolutional layers with 256 channels. 
	
	We use the same multi-scale architecture as Glow to connect the layers, so the number of levels $L$ and the number of steps of each level $K$ will also impact the model's performance. We use Adam~\cite{kingma2014adam} to tune the learning rates, with $\alpha = 0.0002$, $\beta_1 = 0.9$, and $\beta_2 = 0.999$. We set the mini-batch size to be $2$. Based on our empirical results, these settings allow the model to converge quickly. For the experiments on small datasets, i.e., semantic segmentation and image denoising, we run the program for $5 \times 10^4$ iterations to guarantee the algorithms have fully converged. For the experiments on inpainting, the training set is large, so we run the program for $3 \times 10^5$ iterations.
	
	\subsection{Binary Segmentation}
	
	In this set of experiments, we use the Weizmann Horse Image Database~\cite{borenstein2002class}, which contains $328$ images of horses and their segmentation masks indicating whether pixels are part of horses or not. The training set contains 200 images, and the test set contains 128 images. We compare c-Glow with DVN~\cite{gygli2017deep}, NLStruct~\cite{graber2018deep}, and FCN\footnote{We use code from https://github.com/wkentaro/pytorch-fcn.}~\cite{long2015fully}. Since the code for DVN and NLStruct is not available online, we reproduce results of DVN and NLStruct by \citeauthor{gygli2017deep} (\citeyear{gygli2017deep}), and \citeauthor{graber2018deep} (\citeyear{graber2018deep}). We use mean intersection-over-union (IOU) as the metric. We resize the images and masks to be $32 \times 32$,  $64 \times 64$, and $128 \times 128$ pixels. For c-Glow, we follow \citeauthor{kingma2018glow} (\citeyear{kingma2018glow}) to preprocess the masks; we copy each mask three times and tile them together, so $y$ has three channels.  This transformation can improve the model performance. We set $L=3$, $K=8$, $n_c=64$, and $n_w=128$. 
	\begin{table}[htp]
		\begin{center}
			\caption{Binary segmentation results (IOU).}
			\label{tab:table1}
			\begin{tabular}{lllll}
				\toprule
				\textbf{Image Size} & \textbf{c-Glow} & \textbf{FCN} & \textbf{DVN} & \textbf{NLStruct} \\
				\midrule
				$32 \times 32$ & 0.812& 0.558 &0.840 & ---\\
				$64 \times 64$ & 0.852 & 0.701 & --- & 0.752\\
				$128 \times 128$&  0.858 &0.795&---&---\\
				\bottomrule
			\end{tabular}
		\end{center}
		\vspace{-0.1in}
	\end{table}
	
	Table~\ref{tab:table1} lists the results. DVN only has result on $32 \times 32$ images, and NLStruct only has result on $64 \times 64$ images. The NLStruct is tested on a smaller test set with 66 images. In our experiments, we found that the smaller test set does not have significant impact on the IOUs. DVN and NLStruct are deep energy-based models. FCN is a feed-forward deep model specifically designed for semantic segmentation. Energy-based models outperform FCN, because they use energy functions to capture the dependencies among output labels. Specifically, DVN performs the best on $32 \times 32$ images. The papers on DVN and NLStruct do not include results for large images. Thus, we only include small image results for DVN and NLStruct. C-Glow can easily handle larger size structured prediction tasks, e.g., $128 \times 128$ images. Even though c-Glow performs slightly worse than DVN on small images, it significantly outperforms FCN and NLStruct on larger images. The IOUs of c-Glow on larger images are also better than DVN on small images.
	
	\subsection{Multi-class Segmentation}
	
	In this set of experiments, we use the Labeled Faces in the Wild (LFW) dataset~\cite{huang2007unsupervised,kae2013augmenting}. It contains 2,927 images of faces, which are segmented into three classes: face, hair, and background.  We use the same training, validation, and test split as previous works \cite{kae2013augmenting,gygli2017deep},  and super-pixel accuracy (SPA) as our metric. Since c-Glow predicts the pixel-wise label,  we follow previous papers \cite{tsogkas2015deep,gygli2017deep} and use the most frequent label in a super-pixel as its class. We resize the images and masks to be $32 \times 32$,  $64 \times 64$, and $128 \times 128$ pixels. We compare our method with DVN and FCN. For c-Glow, we set $L=4$, $K=8$, $n_c=64$, and $n_w=128$. Note that comparing with binary segmentation experiments, we increase the model size by adding one more level. This is because the LFW dataset is larger and multi-class segmentation is more complicated.
	
	\begin{table}[htp]
		\begin{center}
			\caption{Multi-class segmentation results (SPA).}
			\label{tab:table2}
			\begin{tabular}{llll}
				\toprule
				\textbf{Image Size} & \textbf{c-Glow} & \textbf{FCN} & \textbf{DVN} \\
				\midrule
				$32 \times 32$ & 0.914& 0.745 &0.924  \\
				$64 \times 64$ & 0.931 & 0.792 & ---\\
				$128 \times 128$ & 0.945 &0.951&---\\
				\bottomrule
			\end{tabular}
		\end{center}
	\end{table}
	
	The results are in Table~\ref{tab:table2}. On $32 \times 32$ images, DVN performs the best, but c-Glow is comparable. C-glow performs better than FCN on $64 \times 64$ images, but slightly worse than FCN on $128 \times 128$ images. FCN performs well on large images, but worse than other methods on small images. We attribute this to two reasons. First,  for small images, the input features do not contain enough information. The inferences of c-Glow and DVN combine the features as well as the dependencies among output labels to lead to better results. In contrast, FCN predicts each output independently, so it is not able to capture the relationship among output variables. On larger images, the higher resolution makes segmented regions wider in pixels~\cite{long2015fully,gygli2017deep}, so a feed-forward network that produces coarser and smooth predictions can perform well. C-Glow's performance is stable. Whether on small images or large images, it is able to generate good quality results. Even though it is slightly worse than the best methods on $32 \times 32$ and $128 \times 128$ images, it significantly outperforms FCN on $64 \times 64$ images. Moreover, c-Glow's SPAs are better than DVN on small images.
	
	\subsection{Color Image Denoising}
	
	In this section, we conduct color image denoising on the BSDS500 dataset~\cite{arbelaez2010contour}. We train models on $400$ images and test them on the commonly used $68$ images~\cite{roth2009fields}.  Following previous work \cite{schmidt2014shrinkage}, we crop a $256 \times 256$ region for each image and resize it to $128 \times 128$. We then add Gaussian noise with standard deviation $\sigma = 25$ to each image. We use peak signal-to-noise ratio (PSNR) as our metric, where higher PSNR is better. We compare c-Glow with some state-of-the-art baselines, including BM3D~\cite{dabov2007image}, DnCNN~\cite{zhang2017beyond}, and McWNNM~\cite{xu2017multi}. DnCNN is a deep feed-forward model specifically designed for image denoising. BM3D and McWNNM are traditional non-deep models for image denoising. For c-Glow, we set $L=3$, $K=8$, $n_c=64$, and $n_w=128$. Let $x$ be the clean images and $\hat{x}$ be the noisy images. To train the model, we follow \citeauthor{zhang2017beyond} (\citeyear{zhang2017beyond}) and use $(\hat{x})$ as the input and $\hat{x} - x$ as the output. To denoise the images, we first predict $y^*$ and then compute $(\hat{x} - y^*)$.
	\begin{table}[htp]
		\begin{center}
			\caption{Color image denoising results (PSNR).}
			\label{tab:table3}
			\begin{tabular}{lllll}
				\toprule
				\textbf{c-Glow}& \textbf{McWNNM} & \textbf{BM3D} & \textbf{DnCNN}\\
				\midrule
				27.61 & 25.58 &28.21 & 28.53 \\
				\bottomrule
			\end{tabular}
		\end{center}
	\end{table}
	
	\begin{figure}[htp]
		\centering
		\tiny{Noisy Image\hspace{0.425in} Ground Truth \hspace{0.475in} c-Glow \hspace{0.5in} DnCNN \hspace{0.1in}\phantom{.}}
		\includegraphics[width=1\linewidth]{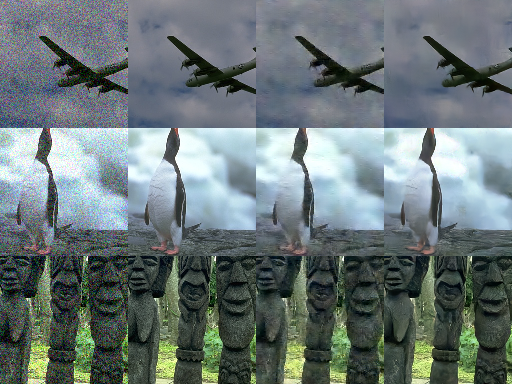}  
		\caption{Example qualitative results.}
		\label{fig:rgb_denoising1}
		\vspace{-0.1in}
	\end{figure}
	
	The PSNR comparisons are in Table~\ref{tab:table3}. C-Glow produces reasonably good results. However, it is worse than DnCNN and BM3D. To further analyze c-Glow's performance, we show qualitative results in Figure~\ref{fig:rgb_denoising1}. One main reason the PSNR of c-Glow is lower than DnCNN is that the images generated by DnCNN are smoother than the images generated by c-Glow. We believe this is caused by one drawback of flow-based models. Flow-based models use squeeze layers to fold input tensors to exploit the local correlation structure of an image. The squeeze layers use a spatial pixel-wise checkerboard mask to split the input tensor, which may cause values of neighbor pixels to vary non-smoothly.

	\subsection{Denoising for Depth Refinement}
	
	In this set of experiments, we use the seven scenes dataset~\cite{newcombe2011kinectfusion}, which contains noisy depth maps of natural scenes. The task is to denoise the depth maps.
	We use the same method as \citeauthor{wang2016proximal} (\citeyear{wang2016proximal}) to process the dataset. We train our model on $200$ images from the Chess scene and test on 5,500 images from other scenes. The images are randomly cropped to $96 \times 128$ pixels. We use PSNR as the metric. We compare c-Glow with ProximalNet~\cite{wang2016proximal}, FilterForest~\cite{ryan2014filter}, and BM3D~\cite{dabov2007image}. For c-Glow, the parameters are set to be $L=3, K=8, n_c=8$, and $n_w=32$. Note that we use smaller conditioning networks for this task, because the images for this task are one-dimensional grayscale images.
	
	We list the metric scores in Table~\ref{tab:table4}. ProximalNet is a deep energy-based structured prediction model, and FilterForest and BM3D are traditional filter-based models. ProximalNet works better than filter-based baselines, and c-Glow gets a slightly better PSNR.
	
	\begin{table}[htbp]
		\begin{center}
			\caption{Depth refinement scores (PSNR).}
			\label{tab:table4}
			\begin{tabular}{llll}
				\toprule
				\textbf{c-Glow} & \textbf{ProximalNet} & \textbf{FilterForest} & \textbf{BM3D}\\
				\midrule
				36.53 & 36.31 & 35.63 & 35.46 \\
				\bottomrule
			\end{tabular}
		\end{center}
		\vspace{-0.3cm}
	\end{table}
	
	\subsection{Image Inpainting}
	
	Inferring parts of images that are censored or occluded requires modeling of the structure of dependencies across pixels. In this set of experiments, we test c-Glow on the task of inpainting censored images from the CelebA dataset~\cite{liu2015faceattributes}, which has around 200,000 images of faces. We randomly select 2,000 images as our test set. We centrally crop the images and resize them to $64 \times 64$ pixels. We use central block masks such that $25\%$ of the pixels are hidden from the input. For c-Glow, we set $L=3, K=8, n_c=64$, and $n_w=128$. For training the model, we set the features $x$ to be the occluded images and the labels $y$ to be the center region that needs to be inpainted. We compare our method with DCGAN inpainting (DCGANi)~\cite{yeh2017semantic}, which is the state-of-the-art deep model for image inpainting. We use PSNR as our metric. 
	
	\begin{table}[htbp]
		\begin{center}
			\caption{Image inpainting scores (PSNR). ``DCGANi-b" represents DCGANi with Poisson blending.}
			\label{tab:table5}
			\begin{tabular}{lll}
				\toprule
				\textbf{c-Glow} & \textbf{DCGANi-b} & \textbf{DCGANi}\\
				\midrule
				24.88 & 23.65 & 22.73\\
				\bottomrule
			\end{tabular}
		\end{center}
	\end{table}
	\vspace{-0.1in}
	
	\begin{figure}[htp]
		\centering
		\tiny{\hspace{-0.1in}Ground Truth\hspace{0.15in} Corrupted Image \hspace{0.2in} DCGANi \hspace{0.25in} DCGANi-b \hspace{0.25in} c-Glow}
		\includegraphics[width=1\linewidth]{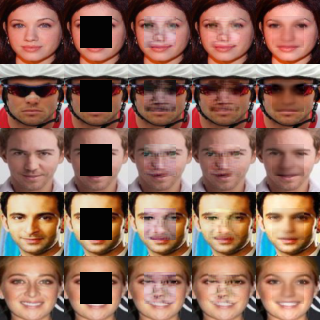}  
		\caption{Sample results of c-Glow and DCGAN inpainting.}
		\label{fig:inp}
	\end{figure}
	
	The PSNR scores are in Table~\ref{tab:table5}. Figure~\ref{fig:inp} contains sample inpainting results. C-Glow outperforms DCGAN inpainting in both the PSNR scores and the quality of generated images. Note that the DCGAN inpainting method largely depends on postprocessing the images with Poisson blending, which can make the color of the inpainted region align with the surrounding pixels. However, the shapes of features like noses and eyes are still not well recovered. Even though the images inpainted by c-Glow are slightly darker than the original images, the shapes of features are well captured.
	
	\subsection{Discussion}
	
	We evaluated c-Glow on five different structured prediction tasks. Two tasks require discrete outputs (binary and multi-class segmentation) while the other three tasks require continuous variables. C-Glow works well on all the tasks and scores comparably to the best method for each task. We compare c-Glow with different baselines for each task, some specifically designed for that task and some that are general deep energy-based models. Our results show that c-Glow outperforms deep energy-based models on many tasks, e.g., scoring higher than DVN and NLStruct on binary segmentation. C-Glow also outperforms some deep models on some tasks, e.g., DCGAN inpainting. However, c-Glow's generated images are not smooth enough, so its PSNR scores are slightly below DnCNN and BM3D for denoising. C-Glow handles these different tasks with the same CN architecture with only slight changes to the size of latent networks, demonstrating c-Glow to be a strong general-purpose model.
	
	\section{Conclusion}
	\label{sec:conclusion}
	
	In this paper, we propose conditional generative flows (c-Glow), which are conditional generative models for structured output learning. The model allows the change-of-variables formula to transform conditional likelihood for high-dimensional variables. We show how to convert the Glow model to a conditional form by incorporating conditioning networks. Our model can train by directly maximizing exact likelihood, so it does not need surrogate objectives or approximate inference. With a learned model, we can efficiently draw conditional samples from the exact learned distribution. Our experiments test c-Glow on five structured prediction tasks, finding that c-Glow generates accurate conditional samples and has predictive abilities comparable to recent deep structured prediction approaches. In the future, we will develop different variants of c-Glow for more complicated tasks in NLP, and test them on different datasets.
	
	\section*{Acknowledgments}
	\label{sec:acknowledgments}
	We thank NVIDIA's GPU Grant Program and Amazon's AWS Cloud Credits for Research program for their support.

	\fontsize{9.4pt}{10.5pt} \selectfont


\newpage

\fontsize{10pt}{11pt} \selectfont
\appendix
\section{Experiment Details}
\label{sec:appendix}
In this section, we introduce more details of our experiments.

\subsection{Network Architectures}
\begin{figure}[!htp]
	\centering
	\begin{subfigure}{0.23\textwidth}
		\centering
		\includegraphics[width=.33\linewidth]{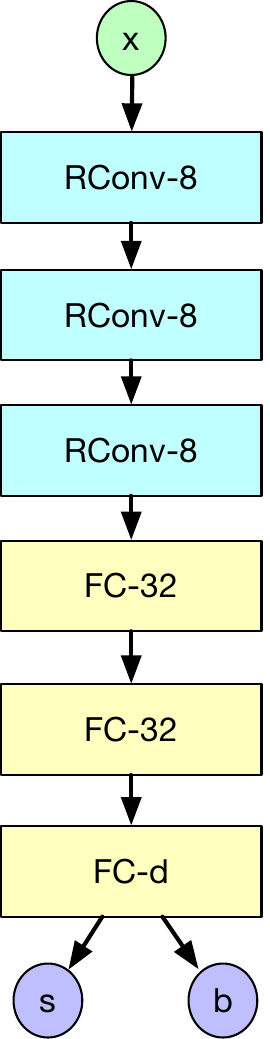}  
		\caption{Conditional Actnorm}
		\label{fig:a}
	\end{subfigure}
	\begin{subfigure}{0.23\textwidth}
		\centering
		\includegraphics[width=.33\linewidth]{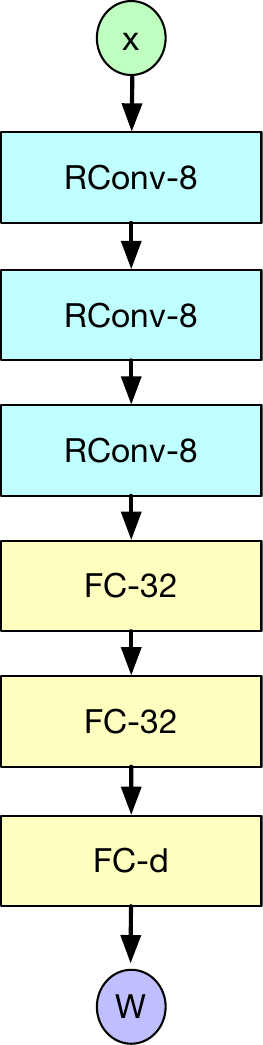}  
		\caption{Conditional 1x1 Convolutional}
		\label{fig:b}
	\end{subfigure}
	\begin{subfigure}{0.5\textwidth}
		\centering
		\includegraphics[width=.23\linewidth]{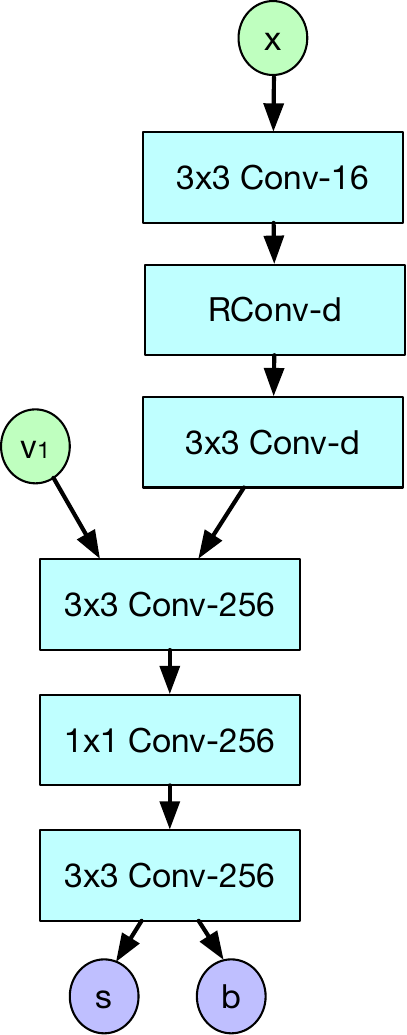}  
		\caption{Conditional Affine}
		\label{fig:b}
	\end{subfigure}
	\caption{The networks we use to generate weights. The component ``3x3 Conv-256" is a convolutional layer, the kernel size is $3 \times 3$, and the number of channels is $256$. The component ``FC-32" is a fully connected layer, and its width is $32$. The parameter $d$  depends on other variable sizes. In the conditional affine layer, $d$ equals the number of channels of $v_1$. In the conditional actnorm layer, $d = 2c$, where $c$ is the size of the scale. In the conditional 1x1 convolutional layer,  $W$ is a $c \times c$ matrix, so $d = c^2$. The ``RConv" component is the convolutional layer for downscaling the input.}
	\label{fig:nn}
\end{figure}

Figure~\ref{fig:nn} illustrates the architectures of conditioning networks that we use in our experiments. For each layer except for the last layer, we use ReLU to activate the output. As in Glow, we use zero initialization for each layer. That is, we initialize the weights of each layer to be zero.
\vspace{0.3in}

\newpage
\subsection{Conditional Likelihoods}
\begin{figure}[htbp]
	\centering
	\begin{subfigure}{0.4\textwidth}
		\centering
		\includegraphics[width=0.7\linewidth]{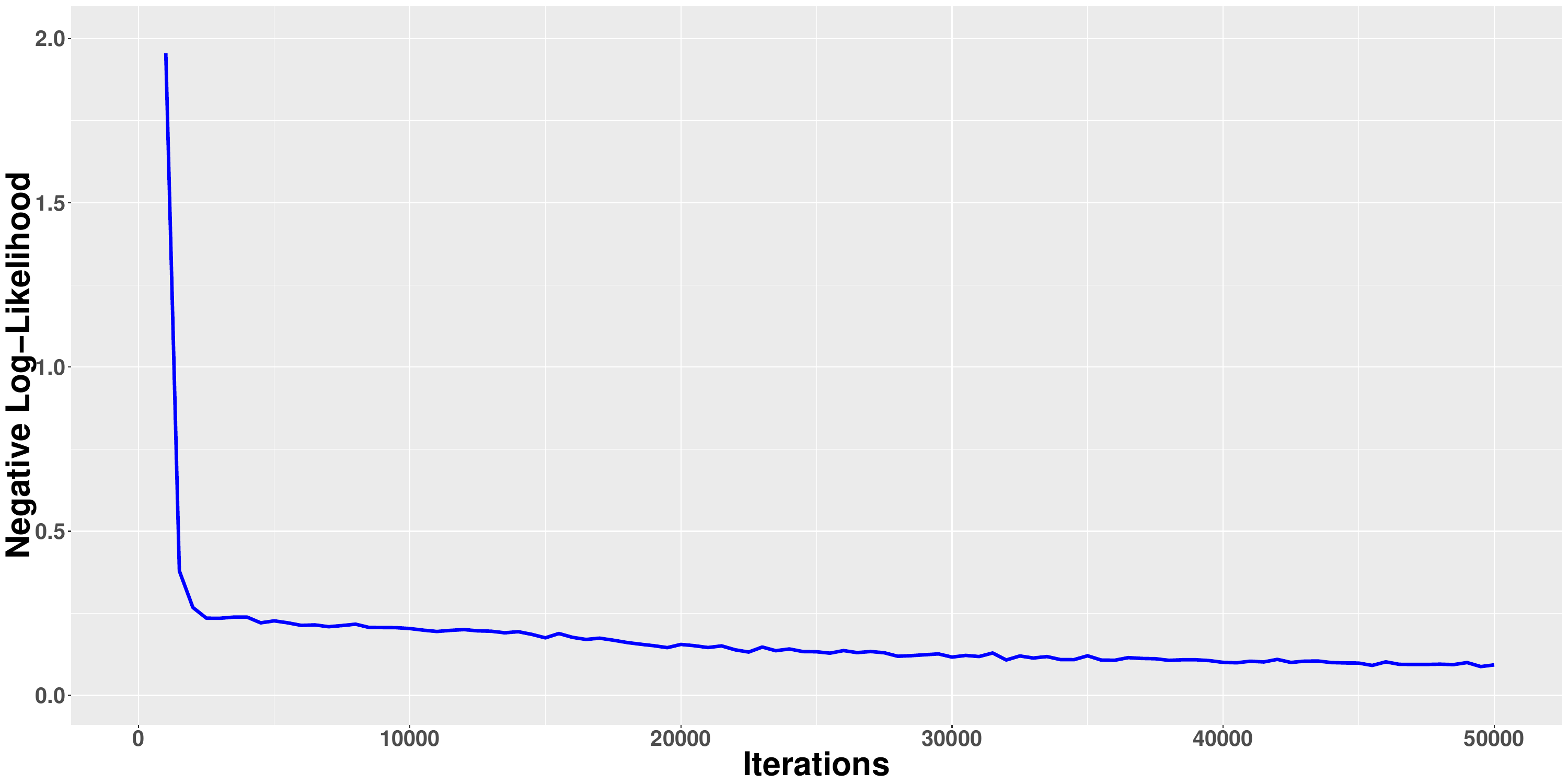}  
		\caption{Binary Segmentation}
		\label{fig:a}
	\end{subfigure}
	\begin{subfigure}{0.4\textwidth}
		\centering
		\includegraphics[width=0.7\linewidth]{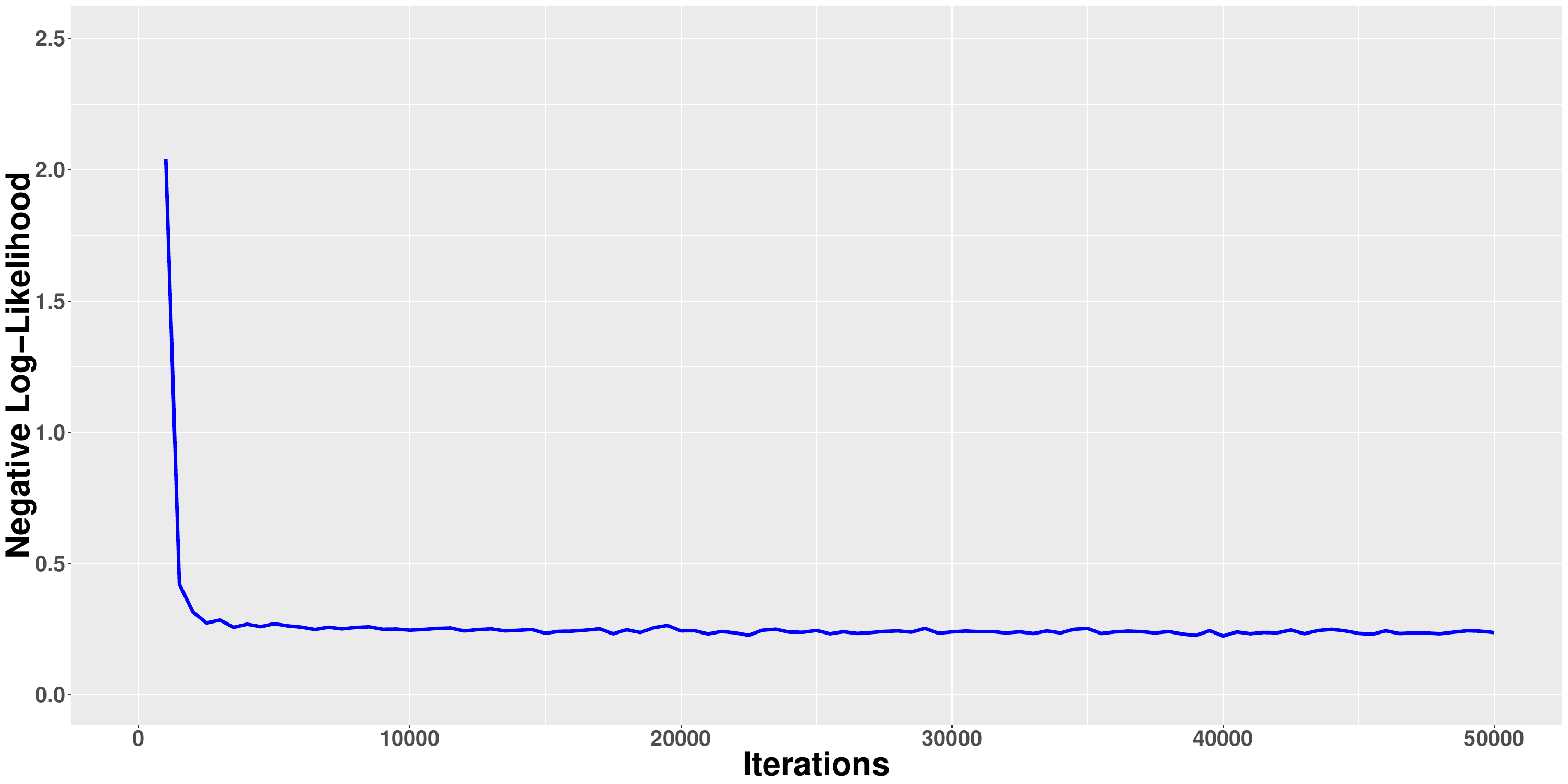}  
		\caption{Multi-class Segmentation}
		\label{fig:a}
	\end{subfigure}
	\begin{subfigure}{0.4\textwidth}
		\centering
		\includegraphics[width=0.7\linewidth]{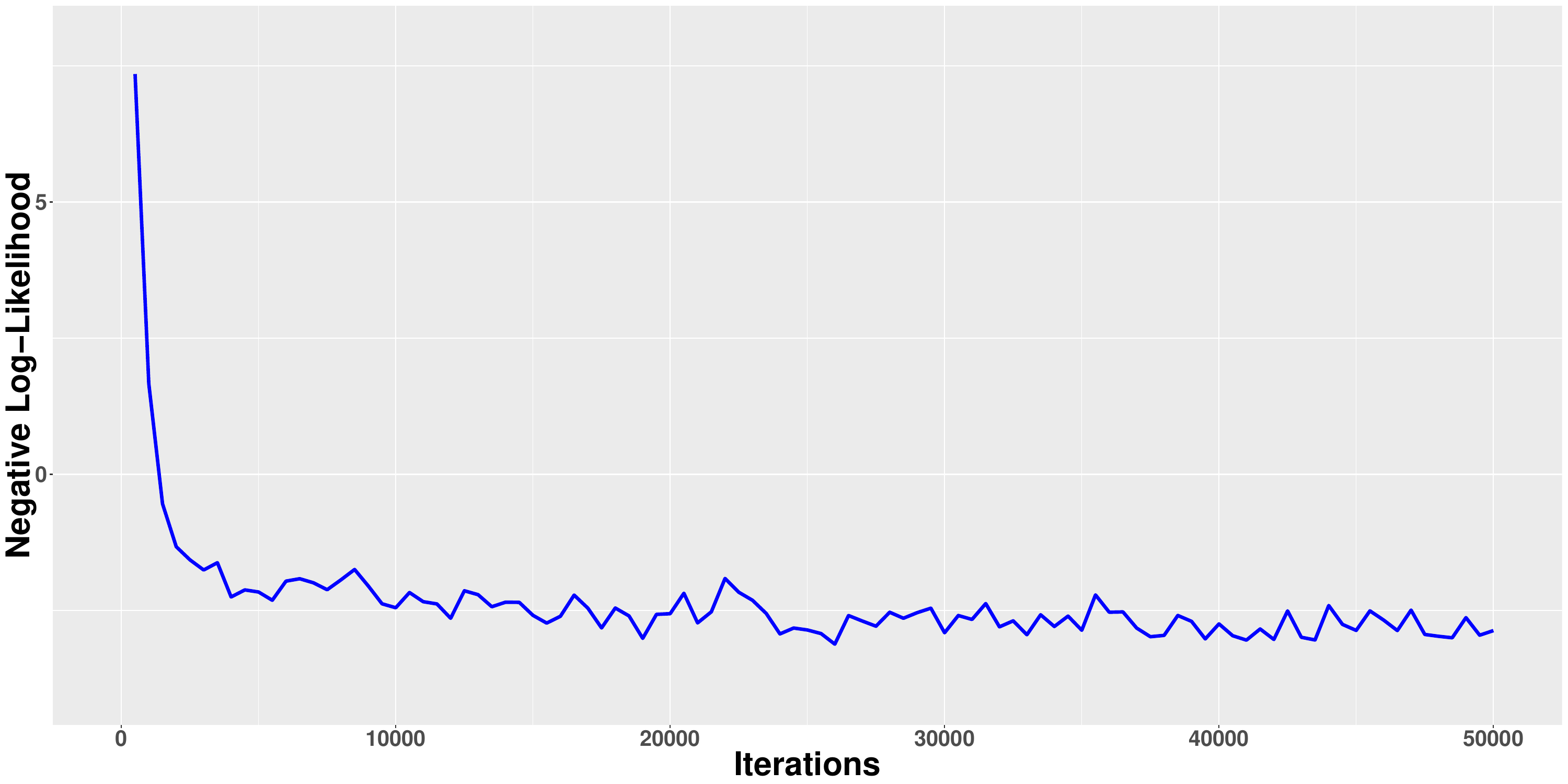}  
		\caption{Color Image Denoising}
		\label{fig:a}
	\end{subfigure}
	\begin{subfigure}{0.4\textwidth}
		\centering
		\includegraphics[width=0.7\linewidth]{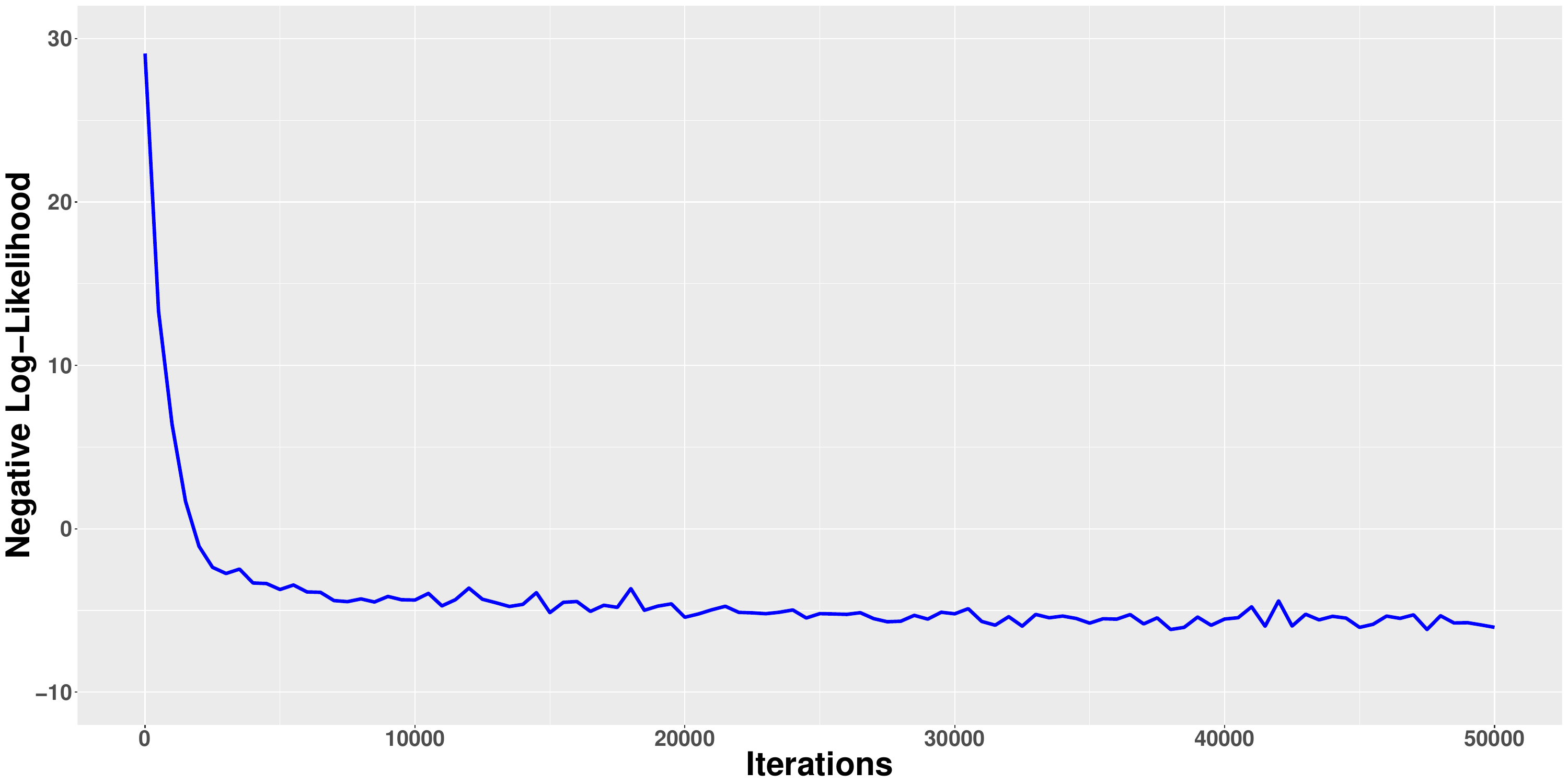}  
		\caption{Depth Refinement}
		\label{fig:b}
	\end{subfigure}
	\begin{subfigure}{0.4\textwidth}
		\centering
		\includegraphics[width=0.7\linewidth]{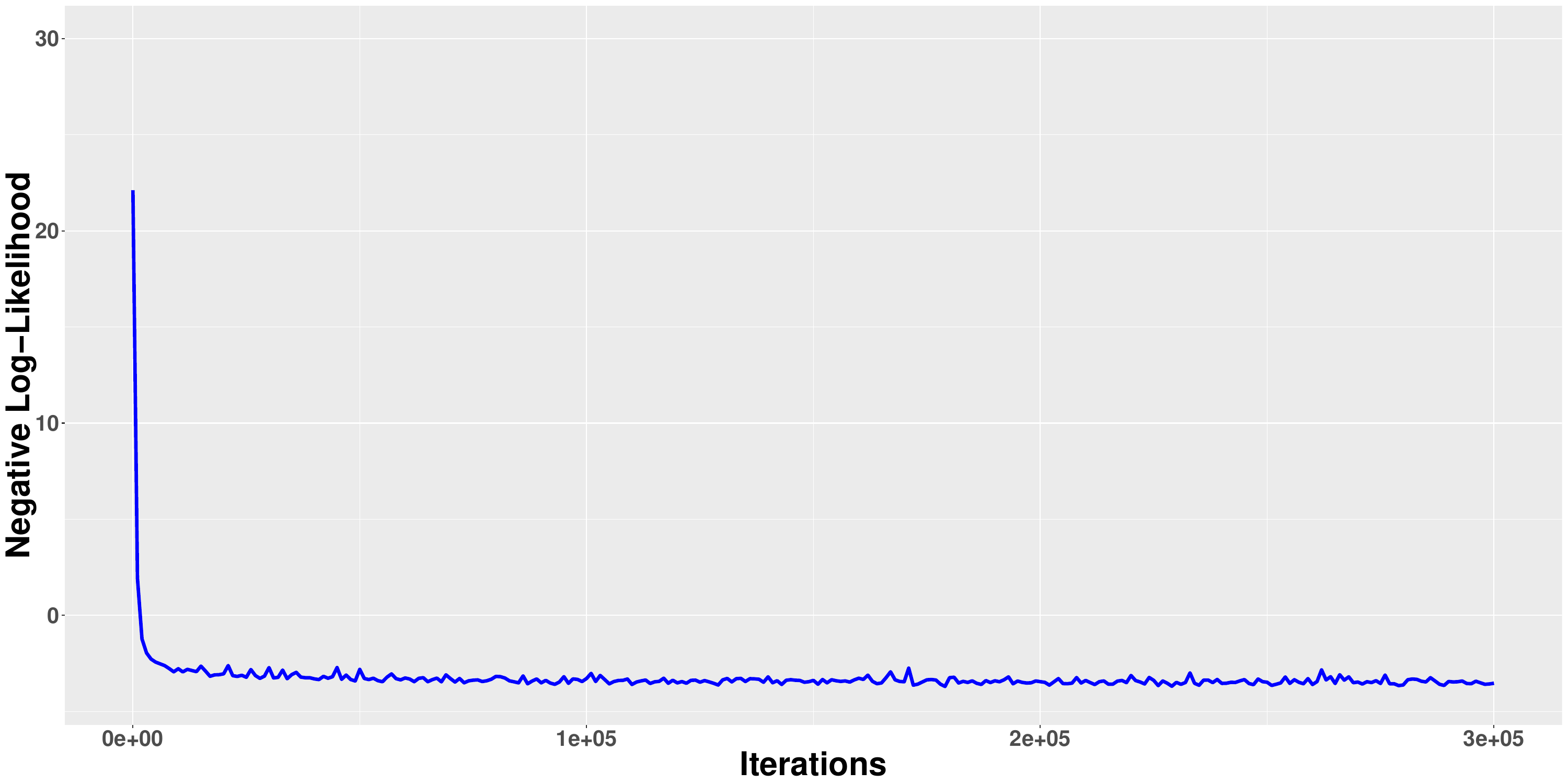}  
		\caption{Image Inpainting}
		\label{fig:c}
	\end{subfigure}
	\caption{Evolution of likelihoods.}
	\label{fig:nll}
\end{figure}

To the best of our knowledge, c-Glow is the first deep structured prediction model whose exact likelihood is tractable. Figure~\ref{fig:nll} plots the evolution of minibatch negative log likelihoods during training. Since c-Glow learns a continuous density, the negative log likelihoods can become negative as the model better fits the data distribution.
\vspace{0.3in}

\newpage
\subsection{Conditional Samples and Predictions}

\begin{figure}[htbp]
	\centering
	\begin{subfigure}[t]{0.5\textwidth}
		\centering
		\includegraphics[width=0.8\linewidth]{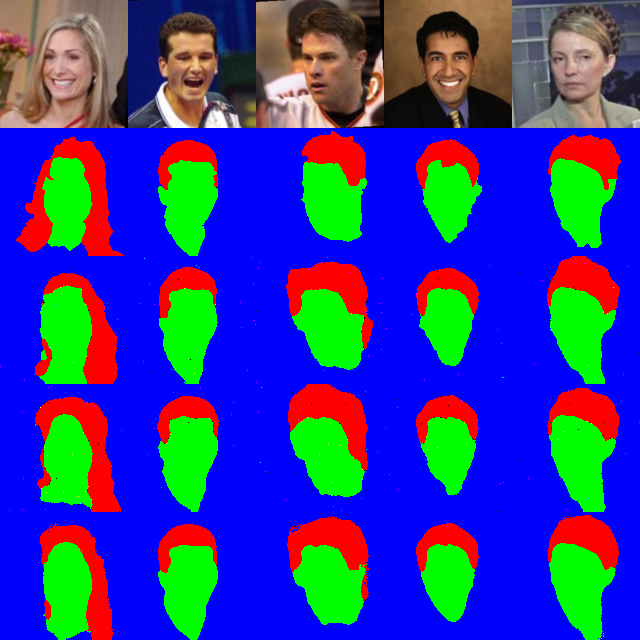}  
		\caption{}
		\label{fig:a}
	\end{subfigure}
	\begin{subfigure}[t]{0.5\textwidth}
		\centering
		\includegraphics[width=0.8\linewidth]{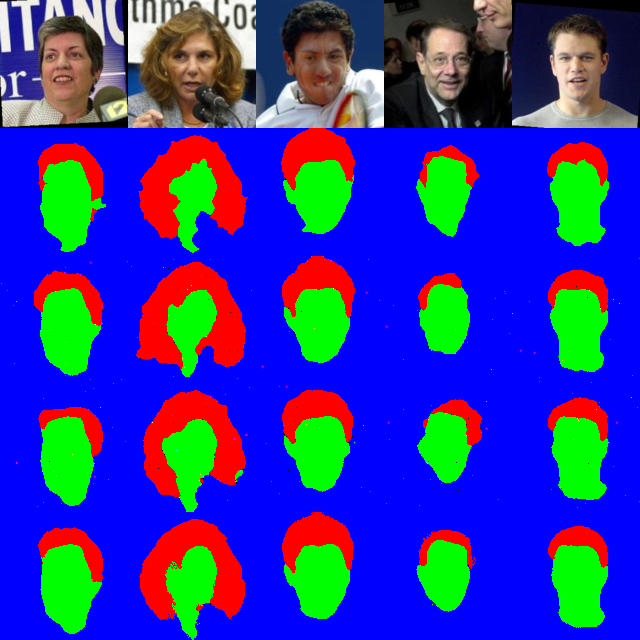}  
		\caption{}
		\label{fig:b}
	\end{subfigure}
	\begin{subfigure}[t]{0.5\textwidth}
		\centering
		\includegraphics[width=0.8\linewidth]{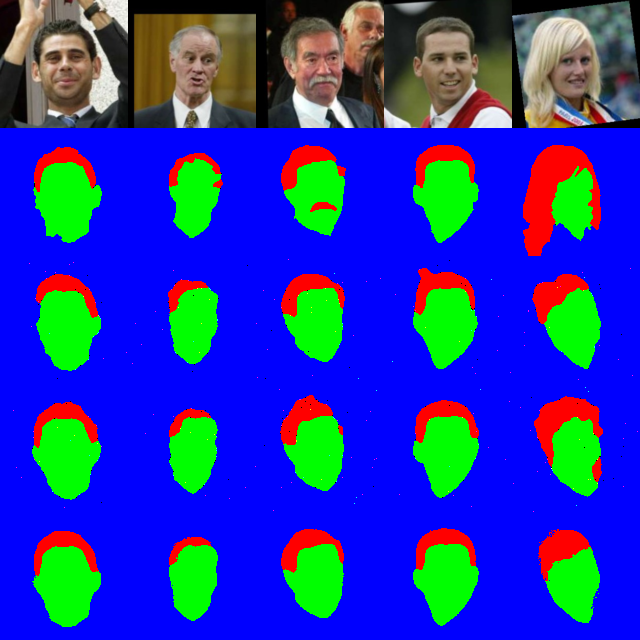}  
		\caption{}
		\label{fig:c}
	\end{subfigure}
	\caption{Conditional samples and predictions on the LFW dataset. The first two rows are input images and ground truth labels, the third and fourth rows are conditional samples, and the last row is predicted labels.}
	\label{fig:m_seg}
\end{figure}

\begin{figure}[htbp]
	\centering
	\begin{subfigure}[t]{0.5\textwidth}
		\centering
		\includegraphics[width=1\linewidth]{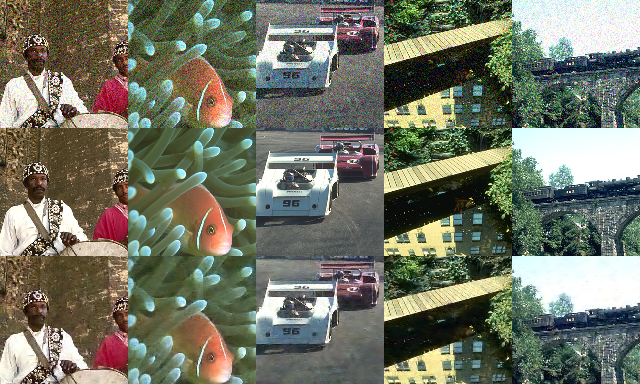}  
		\caption{}
		\label{fig:a}
	\end{subfigure}
	\begin{subfigure}[t]{0.5\textwidth}
		\centering
		\includegraphics[width=1\linewidth]{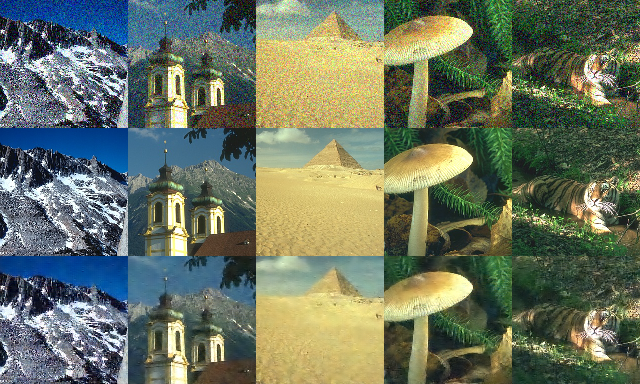}  
		\caption{}
		\label{fig:b}
	\end{subfigure}
	\begin{subfigure}[t]{0.5\textwidth}
		\centering
		\includegraphics[width=1\linewidth]{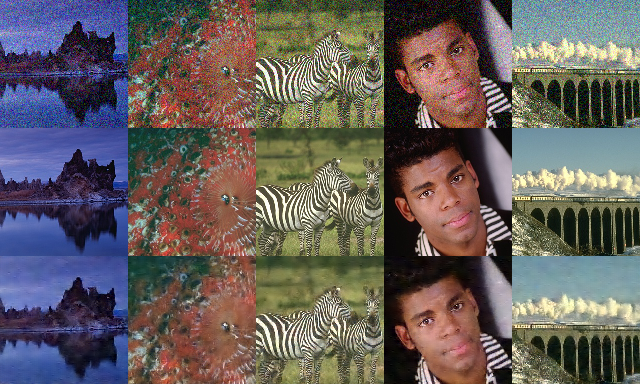}  
		\caption{}
		\label{fig:c}
	\end{subfigure}
	\caption{Conditional samples and predictions on the BSDS dataset. From top to bottom: the noisy images, the clear images, and the denoised images.}
	\label{fig:denoising}
\end{figure}

\begin{figure}[htbp]
	\centering
	\begin{subfigure}[t]{0.5\textwidth}
		\centering
		\includegraphics[width=0.9\linewidth]{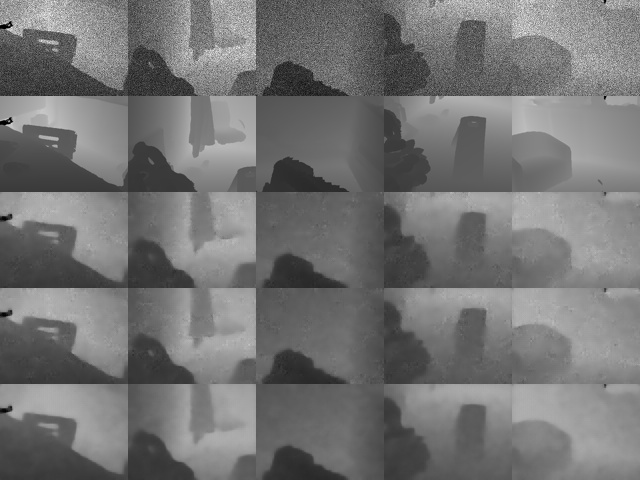}  
		\caption{}
		\label{fig:a}
	\end{subfigure}
	\begin{subfigure}[t]{0.5\textwidth}
		\centering
		\includegraphics[width=0.9\linewidth]{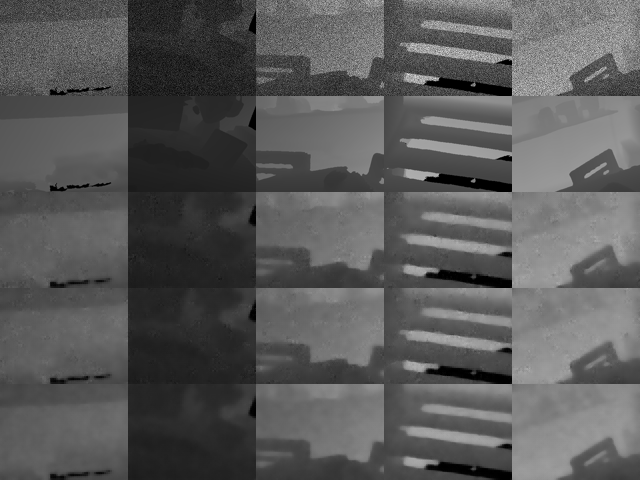}  
		\caption{}
		\label{fig:b}
	\end{subfigure}
	\begin{subfigure}[t]{0.5\textwidth}
		\centering
		\includegraphics[width=0.9\linewidth]{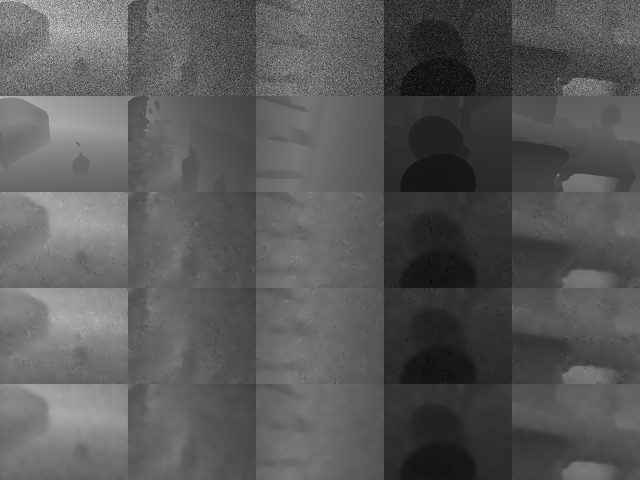}  
		\caption{}
		\label{fig:c}
	\end{subfigure}
	\caption{Conditional samples and predictions on the seven scenes dataset. The first two rows are input images and ground truth labels, the third and fourth rows are conditional samples, and the last row is predicted labels.}
	\label{fig:depth}
\end{figure}

\begin{figure}[hbtp]
	\centering
	\begin{subfigure}[t]{0.5\textwidth}
		\centering
		\includegraphics[width=1\linewidth]{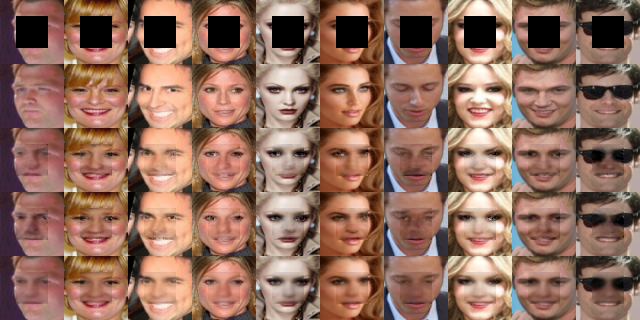}  
		\caption{}
		\label{fig:a}
	\end{subfigure}
	\begin{subfigure}[t]{0.5\textwidth}
		\centering
		\includegraphics[width=1\linewidth]{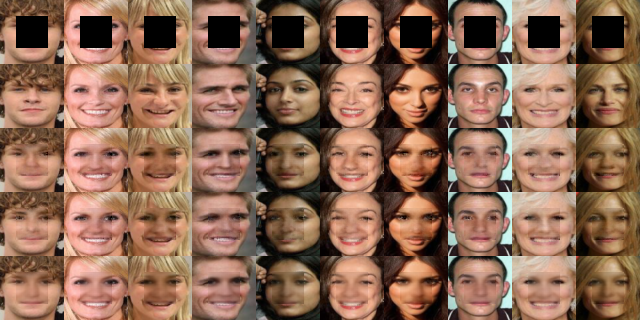}  
		\caption{}
		\label{fig:b}
	\end{subfigure}
	\begin{subfigure}[t]{0.5\textwidth}
		\centering
		\includegraphics[width=1\linewidth]{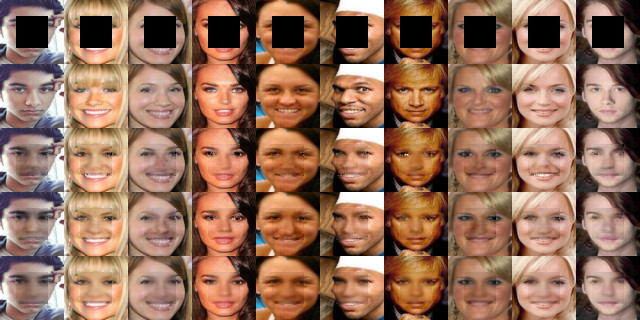}  
		\caption{}
		\label{fig:c}
	\end{subfigure}
	\caption{Inpainting results on the CelebA dataset. The first row is the corrupted input. The second row is the ground truth labels. The third and fourth rows are conditional samples, and the last row is the inpainting results. We use sample average as the final prediction.}
	\label{fig:inpaint}
\end{figure}

One important advantage of our model is the ability to easily generate high quality conditional samples. In this section, we show some conditional samples as well as prediction results in the following figures. Specifically, Figure~\ref{fig:b_seg} shows the binary segmentation results on the Horses dataset, Figure~\ref{fig:m_seg} shows the multi-class segmentation results on the LFW dataset, Figure~\ref{fig:denoising} shows the denoised images on the BSDS dataset, Figure~\ref{fig:depth} shows the refined depth images on the seven scenes dataset, and Figure~\ref{fig:inpaint} shows the impainted images on the CelebA dataset. For each image except for Figure~\ref{fig:denoising}, we show conditional samples in the third and fourth rows. For the experiments of color image denoising, since the conditional samples are just noise added to images, we omit them in Figure~\ref{fig:denoising}. 

For the experiments on semantic segmentation, i.e., Figure~\ref{fig:b_seg} and Figure~\ref{fig:m_seg}, the conditional samples are continuous. However as shown in the figures, the generated continuous values are already close to integral. The conditional samples can reflect the divergence in predictions. For example, in the samples of horses, i.e., Figure~\ref{fig:b_seg}, some sampled horses have different shapes of heads or tails, even though they are conditioned on the same input image. Similar phenomena can also be seen in Figure~\ref{fig:m_seg}. 

For image denoising and depth refinement, the outputs are continuous. As discussed in Section 5, the denoised images are not as smooth as those images obtained by feed-forward networks. However, in Figure~\ref{fig:denoising} and Figure~\ref{fig:depth}, we can see that c-Glow can remove a large amount of noise from the noisy images and perform reasonably well.

For image inpainting, the images inpainted by c-Glow are slightly inconsistent with the surrounding pixels in color, but they are nearly the same. The inpainted images can capture the shapes of features well, even though for faces with sunglasses, c-Glow can also recover the shape and color of the sunglasses. This is why c-Glow can outperform DCGANi. The conditional samples of inpainted images can also reflect the diversity of the model predictions. For example, for some samples of faces, parts such as the mouth and nose are the same, but the eyes stare in different directions.

\begin{figure}[!htbp]
	\centering
	\begin{subfigure}[t]{0.5\textwidth}
		\centering
		\includegraphics[width=0.8\linewidth]{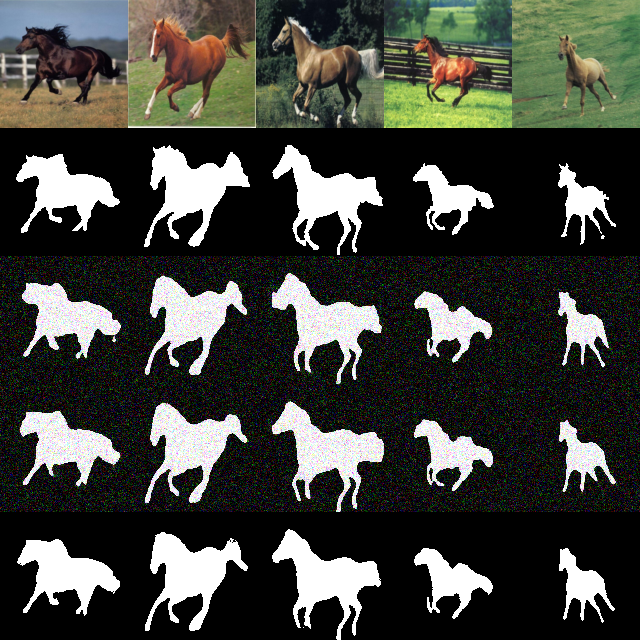}  
		\caption{}
		\label{fig:a}
	\end{subfigure}
	\begin{subfigure}[t]{0.5\textwidth}
		\centering
		\includegraphics[width=0.8\linewidth]{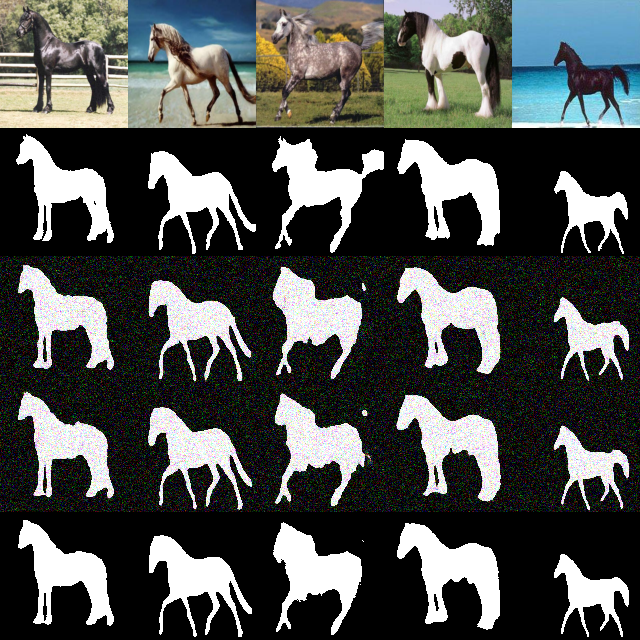}  
		\caption{}
		\label{fig:b}
	\end{subfigure}
	\begin{subfigure}[t]{0.5\textwidth}
		\centering
		\includegraphics[width=0.8\linewidth]{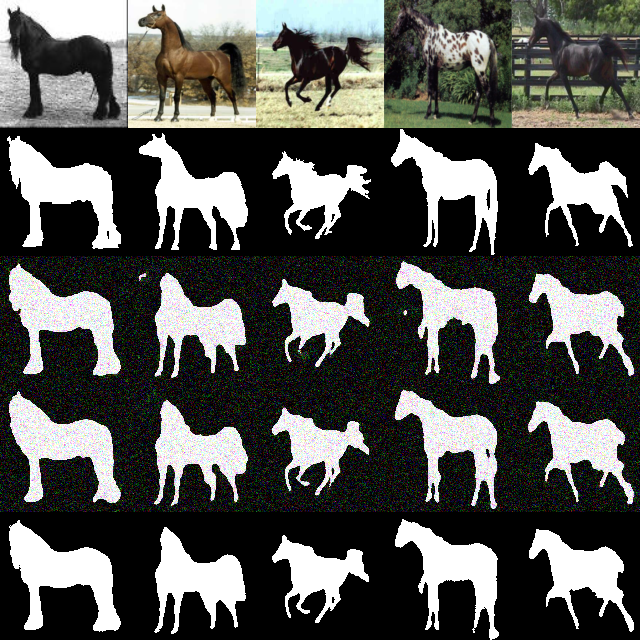}  
		\caption{}
		\label{fig:c}
	\end{subfigure}
	\caption{Conditional samples and predictions on the Horses dataset. The first two rows are input images and ground truth labels, the third and fourth rows are conditional samples, and the last row is predicted labels.}
	\label{fig:b_seg}
\end{figure}

\end{document}